\documentclass[sigconf]{acmart}

\AtBeginDocument{%
  \providecommand\BibTeX{{%
    \normalfont B\kern-0.5em{\scshape i\kern-0.25em b}\kern-0.8em\TeX}}}

\setcopyright{rightsretained}

\usepackage{colortbl}
\usepackage{graphicx} 
\usepackage{url}
\usepackage{stfloats}
\usepackage{tabularx}

\newcommand{\etal}{\textit{et al.\@}}
\newcommand{\eg}{\textit{e.g.,\@}}
\newcommand{\ie}{\textit{i.e.,\@}}
\newcommand{\etc}{\textit{etc.\@}}

\definecolor{Gray}{gray}{0.9}

\usepackage[suppress]{color-edits}
\addauthor{ben}{blue}
\addauthor{andy}{green}
\addauthor{alex}{purple}
\addauthor{MM}{ForestGreen}
\addauthor{ED}{red}

\newcommand*{\templatetitle}[1]{\huge 
#1 
}
\newcommand*{\templatesection}[1]{{\large  
\vspace{4pt}
#1 
\vspace{2pt}

}}
\newcommand*{\templatesubsection}[1]{
\vspace{2pt}
\textbf{#1}:
\vspace{1pt}
}
\newcommand*{\templatetext}[1]{
\small{#1}
}

\newcommand*{\templateexample}[1]{}

\begin{document}

\title{Towards Accountability for Machine Learning Datasets: Practices from Software Engineering and Infrastructure}

\author{Ben Hutchinson, Andrew Smart, Alex Hanna, Emily Denton, Christina Greer, Oddur Kjartansson, Parker Barnes, Margaret Mitchell}
\email{{benhutch,andrewsmart,alexhanna,dentone,ckuhn,oddur,parkerbarnes,mmitchellai}@google.com}

\renewcommand{\shortauthors}{Hutchinson, Smart, Hanna, Denton, Greer, Kjartansson and Mitchell}

\begin{abstract}
Datasets that power machine learning are often used, shared, and re-used with little visibility into the processes of deliberation that led to their creation. As artificial intelligence systems are increasingly used in high-stakes tasks, system development and deployment practices must be adapted to address the very real consequences of how model development data is constructed and used in practice.  This includes  greater transparency about data, and accountability for decisions made when developing it.
In this paper, we introduce a rigorous framework for dataset development transparency that supports decision-making and accountability. The framework uses the cyclical, infrastructural and engineering nature of dataset development to draw on best practices from the software development lifecycle. Each stage of the data development lifecycle yields documents that facilitate improved communication and decision-making, as well as drawing attention to the value and necessity of careful data work. The proposed framework makes visible the often overlooked work and decisions that go into dataset creation, a critical step in closing the accountability gap in artificial intelligence and a critical/necessary resource aligned with recent work on auditing processes.

\end{abstract}

\ccsdesc[500]{Information systems~Data management systems}
\ccsdesc[300]{Computing methodologies~Machine learning}

\keywords{datasets, requirements engineering, machine learning}


\copyrightyear{2021}
\acmYear{2021}
\acmConference[FAccT '21]{Conference on Fairness, Accountability, and Transparency}{March 3--10, 2021}{Virtual Event, Canada}
\acmBooktitle{Conference on Fairness, Accountability, and Transparency (FAccT '21), March 3--10, 2021, Virtual Event, Canada}\acmDOI{10.1145/3442188.3445918}
\acmISBN{978-1-4503-8309-7/21/03}

\maketitle

\section{Introduction}
Machine learning faces a crisis in accountability. Deep learning can match or outperform humans on some tasks \cite{mnih2013playing, liu2019comparison} and is touted as paving the way for achieving human-level artificial general intelligence \cite{arel2012deep}. However the datasets which machine learning (ML) critically depends on---and which frequently contribute to errors---are often poorly documented, poorly maintained, lacking in answerability, and have opaque creation processes. This paper argues that the development of ML  datasets should embrace engineering best practices around visibility and ownership, as a necessary (but not sufficient) requirement for accountability, and as a prerequisite for mitigating harmful impacts.

Despite rapid growth, the disciplines of data-driven decision making---including ML---have come under sustained criticism in recent years due to their tendency to perpetuate and amplify social inequality \cite{barocas2016big,eubanks2018automating}.  Data is frequently identified as a key source of these failures through its role in ``bias-laundering'' \cite{garg2018word, de2019does, richardson2019dirty, gershgorn2018, sambasivan2020}. For example, recent studies have uncovered widespread prevalence of undesirable biases in ML datasets, such as the under-representation of minoritized groups \cite{buolamwini2018gender, shankar2017no, de2019does} and stereotype aligned correlations \cite{zhao2017men, garg2018word, Burns2018, hutchinson2020social}. Datasets also frequently reflect historical  patterns of social injustices, which can subsequently be reproduced by ML systems built from the data. For example, in a recent study examining the datasets underlying predictive policing models deployed in police precincts across the US, the underlying data source was found to reflect racially discriminatory and corrupt policing practices \cite{richardson2019dirty}. The norms and standards of data collection within ML have themselves been subject to critique, with scholars identifying insufficient documentation and transparency regarding processes of dataset construction \cite{geiger2020garbage,gebru2018datasheets, Scheuerman2020}, as well as problematic consent practices \cite{prabhu2020large}. The lack of accountability to datafied and surveilled populations as well as groups impacted by data-driven decisions \cite{Citron2014} has been further critiqued.

Taken together, these objections around data raise a serious challenge to the justifiability of the datasets used by many ML applications. How can AI systems be trusted when the processes that generate the their development data are so poorly understood? If ML is to survive this crisis in a responsible fashion, it must adopt visibility practices that enable accountability and responsibility throughout the data development lifecycle.

This paper takes a step in this direction. We argue for a rigorous and visible dataset creation process that is based in foundational ontological thinking about what kind of ``things'' datasets are. We posit that datasets constitute a form of \textit{technical infrastructure}, in the sense that they are enabling and necessary components of technical systems, with which they interact through technical interfaces. Datasets are also \textit{infrastructure} in the infrastructure studies sense, in that they arrange a way to view and structure knowledge of the world \cite{bowker2000sorting, larkin2013politics}. We show how framing datasets as infrastructure helps to shed light on many of the social and technical phenomena that we see around data; for example, datasets are critical components of ML development but their existence and the conditions of their creation are often invisible and hence undervalued. Further, given their nature as technical infrastructure, the processes that give birth to datasets are best understood as fundamentally goal-driven engineering. We argue that debates around ML data are sometimes complicated by confusions around the epistemological goals of engineering. This is in part due to computer science's own methodological complexity, using varying techniques of knowledge validation that include formal proof, automated testing of both deterministic and stochastic forms, user testing, as well as techniques aimed at providing explainability, traceability and interpretability.  


This paper explores in detail a number of the practices that need to be adopted to mitigate dataset risks. These practices in turn require adopting a deliberative and intentional methodology, rather than the \textit{post hoc} justifications that are sometimes observed when datasets are developed hastily or opportunistically. From an observer's perspective, these practices replace \textit{abductive} and \textit{a posteriori} reasoning about dataset provenance with careful documentation, facilitating reviews and audits \cite{raji2020closing}. They help us to avoid the trap of being blindly ``data-driven'' by poorly-understood datasets---mistaking taking a back seat with ``empirical rigour''---when it is in fact humans who we need to be clearly in the driver's seat in ML development, setting goals, direction and strategy with deliberation and intentionality. These practices thus provide bounds on accountability both in dataset creation, and also of the ML systems that depend on those datasets as infrastructure. 

\begin{figure}
\begin{center}
\includegraphics[scale=0.45]{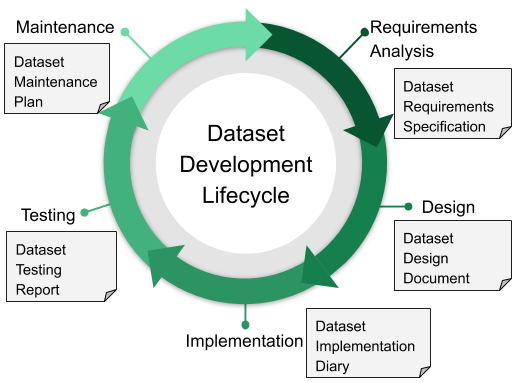}
\end{center}
\caption{The Dataset Development Lifecycle requires documentation for each stage. See Table~\ref{tab:stages} for descriptions of each stage, and Table~\ref{tab:documents} for document types.}
\label{fig:lifecycle}
\end{figure}

We stress the importance of establishing a \textbf{non-linear cycle of dataset development}, with analysis, design and evaluation central to the process (see Figure~\ref{fig:lifecycle}).
Building off this lifecycle, key practices which we will cover include:

\begin{itemize}
\item  \textbf{Documentation:} a model of \textbf{documentation practices} throughout the dataset development lifecycle, drawing on software lifecycle practices (see Table~\ref{tab:documents} and Section~\ref{sec:documentation}); 
\item  \textbf{Oversight:} \textbf{diverse oversight processes}, including audits and reviews, which leverage these practices (see Section~\ref{sec:audits});
\item \textbf{Maintenance:} \textbf{robust maintenance mechanisms}, including those for addressing technical debt, correcting errors, postmortems and adapting to changing contexts (see Sections~\ref{sec:maintenance} and \ref{sec:welfares}).
\end{itemize}

The structure of this paper is as follows. We first make the case for datasets being a form of technical infrastructure (Section~\ref{sec:infrastructure}). Then, by considering the epistemologies, cultures and methods of engineering, we argue that dataset development is essentially a form of engineering practice (Section~\ref{sec:engineering}). Building on this, we show how software engineering and infrastructure practices provide lessons for dataset accountability (Section~\ref{sec:towards}), including a model for dataset development documentation. We then discuss the implications of this model for broader systems and ecologies in which datasets are maintained and used, proposing research directions and argue for cultural shifts (Section~\ref{sec:discussion}).

\begin{table*}
\begin{tabularx}{\textwidth}{p{0.19\textwidth}p{0.59\textwidth}p{0.17\textwidth}}
\hline
\sc Dataset Account 
&\sc Questions Answered by the Document
&\sc Key Roles
\\\hline
\mbox{Dataset Requirements} Specification
&Is data needed? By whom? What are its intended uses? What properties should it have? Do the uses necessitate constraints on collection and/or annotation?
&Requirements owner; stakeholder; reviewer
\\\hline
Dataset Design Document
&Is a new dataset needed (do existing datasets not meet requirements)? How will requirements be operationalized? What tradeoffs \& assumptions are being made?
&Design owner; domain expert; reviewer
\\\hline
Dataset Implementation Diary
&How were the design decisions implemented? Why were they done this way? What unforeseen circumstances arose when implementing the design?
&Implementation owner;
data creator/labeler
\\\hline
Dataset Testing Report
&Should I use the dataset? Does the dataset meets its requirements? Is the dataset safe to use? Is the dataset likely to have previously unforeseen consequences?
&Data scientist; adversarial tester
\\\hline
Dataset Maintenance Plan
& How will data staleness be detected and fixed? How will errors be fixed? How will affordances for manual interventions be provided?
&Maintainer; funder; bug filer; data contester
\\
\hline
 \end{tabularx} 
\caption{Critical document types for accountable dataset development. Each one is directly analogous to documentation types produced by the Software Development Lifecycle.}
\label{tab:documents}
\end{table*}

\section{Datasets as Infrastructure}
\label{sec:infrastructure}
ML fairness practitioners have identified collection and curation of representative data as the most common way to mitigate model biases \cite{holstein2019improving}, echoing the broader tendency within ML communities to valorize larger datasets as necessarily better---exemplified by \cite{halevy2009unreasonable} and subsequently in \cite{krause2015fine, sun2017revisiting}. However, datasets are not merely a collection of “facts” to be discovered, but rather construct a particular kind of knowledge (see for example \cite{latour2013laboratory}). Given the centrality of data to the field of ML and its criticality to the fairness of ML-based systems, it is useful to consider two questions, one ontological and one explanatory: \textit{what kinds of things are datasets?} and \textit{why do datasets so often seem to have problems?} 

The answers to both these questions are intertwined, and are inseparable from the processes of dataset creation and use. Being essential to ML, datasets have all the classic properties of IT infrastructure: they are shared information delivery bases with well-defined architectures and standardized interfaces, creating value by enabling rapid iteration and new forms of processes, from which their ultimate value is derived \cite{bharadwaj2000resource, ross1996develop}. The dependency is in fact so close that in common parlance the ML model and its training data are sometimes used interchangeably, for example biases in models are frequently seemingly identified with training data bias, or used to abductively infer biases in the training data (\eg{} \cite{davidson2019racial, jung2019earlier, babaeianjelodar2020quantifying}).

Despite the critical role data plays, the systemic devaluation of data work is in plain view. Enormous conferences, prizes, and adulation await new algorithmic achievements on ``benchmark'' datasets (seen as fixed resources, or \textit{data} in the original Latin sense \cite{furner2016data}). In contrast, datasets, like other infrastructures, are often taken for granted after they are built, and fade into the background until they break down \cite{bowker2000sorting, denton2020bringing}. Often playing the role of commons \cite{frischmann2012infrastructure}, benchmark datasets exhibit ``comedy of the commons'' whereby usage of a dataset increases its value, due to the comparisons enabled by its role as a measuring instrument \cite{welty2019metrology}, while the patient and careful work of caring for datasets (including the collection, cleaning, and annotation of data) receives few, if any, accolades in the mainstream field. Since it is often an unpaid burden on top of existing workloads \cite{sambasivan2020, taylor2018automation, irani2013turkopticon}, careful dataset work is unlikely to occur at all. Publications focusing solely on datasets rarely make it through the review process of top-tier venues \cite{heinzerling2020nlp} and---perhaps due to the imperative to squeeze dataset details into methods-oriented papers---critical dataset design decisions are frequently left underspecified (e.g. \cite{geiger2020garbage}). Despite the foundational role data place in ML research and development, data occupies little to no space within predominant ML textbooks and data is not recognized as an area of ML specialization \cite{jo2020lessons}. ML data are also not version-controlled, held at institutional repositories, or given DOIs; and data annotation work is given short shrift in discourses about model building and development.

Thus, it is not solely the invisibility of the dataset to end users (for the learning algorithm is also invisible) that alone explains the problems we see in datasets, but also the devaluing of the processes of dataset creation. And here a vicious circle enters the picture. Because benchmark datasets shift the field's attention to differences between learning algorithms, as measured by narrow benchmark metrics, we see an emphasis on explaining and documenting those differences. In contrast, the benchmark dataset, now a fixed resource for measurement and competition, has little need for explanation or documentation. For datasets, poor valuation leads to poor documentation, which in turn leads to even poorer valuation as the ``fixed'' becomes the ``taken for granted''. An ML dataset, after its release, quickly becomes a ``black box" of scientific inquiry, but possibly without facing requisite ``trials of strength" that technoscience tools in other domains may encounter \cite{latour1987science}.

In many respects, then, dataset development shares inherent similarities with work on other forms of technical infrastructure. It is deep in the training of computer programmers to take infrastructure for granted; indeed, the use of abstractions in software libraries encourages this (the ``portability trap'' of \cite{selbst2019fairness}). Although software developers and model developers both need to ``get things done'' by working with abstractions, they also need processes of justifying and/or fixing what's behind the infrastructural curtains, lest abstractions launder unjustified assumptions. We see that work on computing technical infrastructure faces similar challenges with respect to valuation and appreciation. While product software is directly identified with the ``product'', the networks, datacenters, release management tools, and testing frameworks that are critical to deployment remain (when successful) invisible and taken for granted, or (when not) blamed for failures. Corporate value is assigned to the product software---and measured through clicks, views, \etc{}---while improvements to shared and distributed technical infrastructure are much harder to assign value to.

This extended comparison with infrastructure highlights patterns of valuation that are operative in both ML and engineering domains. By considering the ways in which practices of software infrastructure have fought against devaluation, we see opportunities to increase the valuation of dataset labor. Improvements on many fronts are needed, including changing the practices of academia to view data as the first-class infrastructure of ML: data should be critical to the ML curriculum; papers about data should be recognized and rewarded in ML fora; data should be interrogated rather than taken for granted. A further, but critical, change involves reversing the vicious circle described above, and through robust dataset documentation practices bring more attention to the processes and value of dataset labor. In Section~\ref{sec:documentation} we discuss in detail such documentation processes, but first we explore the dataset development process in more detail.

\section{Dataset Development as Engineering}
\label{sec:engineering}
The invisibility of dataset processes described in the prior section leads to a lack of critical examination of datasets used to build models. Are researchers and engineers justified in trusting the datasets that they use to build their models? Are datasets fit for their intended purposes? Do the datasets contain hidden hazards that can make models biased or discriminatory? What are the human biases that have been built into the datasets? (And what are the implications if your model performs well on a biased dataset?) Understanding the processes by which datasets are conceived and constructed is central to these questions. Specifically, in the systems of dataset creation, what are the processes, how are these processes documented, and \textit{who is responsible for what}? Even more fundamentally, what are the cultures of ML data work that enable these systems? In this section we make the case that dataset development systems are essentially engineering systems, in their goals, practices, solutions and cultures. By considering the nature of engineering , we identify five arguments that dataset development is foundationally a type of engineering.
\subsection{Datasets provide ``knowledge-how''}
What distinguishes engineering from other technical and empirical disciplines, and especially from scientific disciplines typified by the physical sciences? Despite widespread assumptions to the contrary, engineering is not simply applied science, for ``while science aims for knowledge, engineering aims for useful change'' \cite{bulleit2015philosophy}. The distinction is knowing how to accomplish specified goals, in contrast to knowing that something is true; in other words, while science aims for \textit{knowledge-that}, engineering aims for \textit{knowledge-how} \cite{ryle1945knowing}. As such, science is typically concerned with explanation, certainty, universality, abstractness and theory; while engineering is characterized by contingency, probability, particularity and concreteness \cite{goldman2004we}.  The two are interdependent: engineering can be in support of science (\eg{} datasets can be used towards scientific goals); and modern science is heavily dependent on engineering \cite{brooks1994relationship}, and on big data \cite{leonelli2020scientific}. However, due to their infrastructural roles (see Section 2), the development of datasets is fundamentally a form of engineering. Datasets are designed for specific purposes, and their usefulness must be judged with regard to that purpose \cite{xiao2010corpus}. As with engineering, good practice when developing datasets demands careful attention to relevant science (including social sciences and engineering sciences). Indeed, the consultation of relevant science is a cornerstone of ideal engineering.
\subsection{Dataset practices are political}
Given that engineering knowledge is \textit{knowledge-how}, it is critical to dispel the myth (discussed in \cite{green2020data}) that engineering practice is a ``purely technical'' matter. Engineering artefacts are imbued with politics via their roles in sociopolitical systems \cite{winner1980artifacts,goldman2004we}, and so too are  ML datasets inseparable from politics \cite{crawford2014big, prabhu2020large, denton2020bringing}. To be ``just an engineer'' is to acknowledge one's role, stakes, and domains of expertise within a sociotechnical system, while also acknowledging the roles, stakes, and expertise of others. It requires accountability for enabling objectives, even if those objectives are not one's own. Within information science disciplines, engineering roles are  filled by data experts, and Ben Green has suggested a four-stage path for how data experts can meaningfully engage with society \cite{green2020data}: becoming interested in addressing social issues, recognizing the underlying politics, redirecting existing methods, and developing practices and methods for working in partnership with communities. Critically, the second stage requires that skills in critical reflection are as essential to dataset development as knowledge of statistics. This aligns with the prescriptions of Neff \etal{}, who ask data scientists to consider several interpretative and value-laden critiques of their data while working with it \cite{neff2017critique}.
\subsection{Datasets are models of reality}
Central to engineering is the concept of \textit{engineering models}, which are representations of reality used for design and testing prior to real-world deployment \cite{ bulleit2015philosophy}. Engineering models are abstracted simplifications of reality, necessarily incomplete, emphasizing some features of reality while deemphasizing or overlooking others \cite{weisberg2012simulation}; they are plausible aids but ``potentially fallible'' \cite{koen2003discussion}.  ML datasets are in fact prototypical engineering models, representations of facts about the world that cannot be experienced directly, nor often replicated. Like other engineering models, datasets can be developed with a number of different end goals including explanation, development of intuition, instruction, prediction, design, evaluation and experimentation \cite{alvi2013engineers}, and their goals dictate priorities and tradeoffs between capturing different dimensions of reality. Uncertainty is a fundamental feature of all models, including datasets \cite{bulleit2015philosophy}, and robust testing and calibration is required to avoid brittle and unsafe conclusions. Bulleit \etal{} report that in engineering disciplines ``designers rarely get feedback on how well their models represent the specific full-scale projects that they have designed'' \cite{bulleit2015philosophy}, and the infrastructural properties of data discussed above suggest the same is true of ML dataset workers.

\subsection{Dataset development is non-linear}\label{sec:non-linear}
Engineering problems ``tend to be ill-structured” \cite{simon1973structure} or even ``wicked” \cite{rittel1973dilemmas}, meaning they defy deductive reasoning and definitive answers. Instead, engineering involves reconceptualizing a complex situation to facilitate analysis; with feedback loops impacting problem (re-)definition (see \cite{passi2019problem}), rather than focusing narrowly on problem solution. As goal-oriented models of reality, dataset development requires iteration and feedback loops due to changes in stakeholders and goals, as design assumptions become invalidated or better alternatives become available, and fundamentally as reality itself changes. Western culture has devalued engineering  with respect to science (albeit not social science)  precisely because of engineering's context sensitivity, pluralism, and temporal instability \cite{goldman2004we}, mirroring data work's devaluation with respect to ML model work (Section~\ref{sec:infrastructure}).

\subsection{Cultures of dataset work mirror those of software}
Computer programming emerged from an intersection of multiple academic disciplines, and the intrinsic cultures of these fields informed the norms and attitudes that persist in software engineering today \cite{petricek2019cultures}. The rise of ML and its big, shared datasets are subject to some of the same cultural pressures and fault lines; their solutions to collective coordination and communication challenges giving rise to similar cultural norms. Indeed, the four cultures of programming identified by \cite{petricek2019cultures} are evident when looking at the development of ML models and datasets:

\textbf{Mathematical culture}: This is evident in the focus on mathematical and statistical methods at the heart of ML, the value given to formal proofs, and the tendency of practitioners to focus on tractable data problems such as agreement and reliability, rather than more complicated data questions such as external validity (\eg{} whether arrests represent crimes) \cite{jacobs2019measurement,welty2019metrology}.

\textbf{Hacker culture}: This culture is expressed in ML via a focus on individual contributions by ``superstars'' \cite{greene2020}, ``savants'' \cite{ramkilowan2018} and saviours  with brilliant innovative insights. Hacker culture emphasizes moving fast and debugging as errors arise \cite{lavin2020} (neural network interpretability methods \cite{chakraborty2017interpretability} are the stack trace debuggers of ML). This culture devalues the incremental and cooperative care required to create high quality datasets, and has been implicated in the discriminatory effects of technological systems.

\textbf{Managerial culture}: Appreciating software development as inherently sociotechnical, this culture focuses on team structures emphasizing responsibility, shared documentation and auditable processes \cite{gebru2018datasheets, mitchell2019model, raji2020closing}, and advancing these practices for ML data is a core goal of this paper.

\textbf{``Engineering'' culture}: Dominated by practical concerns, and accepting the impossibility of formal guarantees, engineering cultures of software development focuses on reliable tools and practical methods. Confidence in systems is gained not by proof or debugging, but by error measurement and error handling.  In ML data, the emphasis on shared benchmark datasets  is a cornerstone of ML engineering culture \cite{welty2019metrology}, and ``Humans in the Loop" in ML deployments (as a form of robust error handling) also fall under this culture.

These correspondences between the cultures, goals and methods of datasets and engineering point the way forwards. In the following section we explore how best practices in software development can be adapted to improving the visibility and quality of dataset development.

\begin{table*}[bth]
\begin{tabular}{p{1.8cm}p{7cm}p{8.2cm}}
\toprule
\sc Barrier
& \sc Dataset Concerns
& \sc Proposals for Mitigating
\\\midrule
\mbox{Problem of} many hands
& Datasets teams may have indeterminate structure and responsibilities; data may be sourced from third parties, or  ``from the wild''; data labeling teams may be unknown    
& Factor dataset development into discrete stages; assign clear responsibilities to the owner of each stage; robust documentation serving as boundary objects 
\\\midrule
Unintended artifacts
& Intended properties of dataset may not be explicit; datasets may contain unforeseen problems 
& Consult diverse stakeholders; document intentions explicitly; assume data is ``guilty until proven innocent''; adversarial testing
\\\midrule
\mbox{Computer as} scapegoat
& ``Data as scapegoat''; sourcing data from logs of applications; sourcing data from generative models
& Require visibility into conditions of dataset creation; document and justify decisions to use data from system logs; understand and document properties of generative models
\\\bottomrule
 \end{tabular} 
\caption{Three of Nissenbaum's barriers to accountability \cite{nissenbaum1996accountability}, specific data concerns, and proposals in this paper.
}
\label{tab:nissenbaum}
\end{table*}

\section{Towards Accountability for Datasets}
\label{sec:towards}


Accountability has been described as ``fundamentally about answerability of actors for outcomes'' \cite{kohli2018translation}.
Critical to this question is what information is known by whom, and how it is used. In fact, the sharing of information is the first key phase of mechanisms of accountability \cite{wieringa2020}. The second phase involves deliberation and discussion by the ``forum'', as well as requests for more information. During the third and final phase,  the forum imposes consequences. In this section we focus our attention on the first, critical, information phase, and ask two questions: 1) \textit{which actors} should provide information about datasets? and 2) \textit{what information} should these actors provide to enable meaningful discussion and consequences?

The information that is shared as a necessary (but not sufficient) precondition for accountability is referred to technically as \textit{accounts}. The recording of dataset accounts is at its most fundamental a question of bookkeeping, but the details are critical: which books should be kept, what are their stories, and who are their authors? The answers lie in the closely related goals of traceability, attributability, responsibility, accountability, and answerability, the distinctions between which have been the subject of much discussion in Western philosophy. Judgements of fault can be on multiple dimensions of responsibility,\eg{} under one account: ``attributability-responsibility'' concerns to an agent’s character, ``accountability-responsibility'' refers to an agent’s regard for others, and ``answerability-responsibility'' is about an agent’s evaluative judgments \cite{shoemaker2011attributability, shoemaker2015}. In this paper, we have little to say about judgments of character, other than calling for transparency about which actors bear attributality-responsibility. (Reliability is clearly a concern here: does this institution/individual have a history of developing trustworthy and responsible artefacts, cf. \cite{smart2020reliabilism}?) However, below we will detail how both regard for others and evaluative decisions are things for which accounts can and should be kept when developing datasets.

Helen Nissenbaum \cite{nissenbaum1996accountability} describes barriers to accountability of computer systems which are all directly applicable to datasets. Table~\ref{tab:nissenbaum} summarizes how her concerns relate to datasets, as well as the mitigating proposals on which we will elaborate below.\footnote{Nissenbaum's fourth concern is about liability, which is beyond the scope of this paper.}
Our analysis and proposals follow \cite{wieringa2020} and \cite{kohli2018translation} in using temporal situatedness as a key organizing factor. At the highest level, what are the dataset accounts that should be kept before (\textit{ex ante}), during (\textit{in medias res}) and after (\textit{ex post}) the collection and labeling of data? Based on the robust parallels with software engineering argued for in Section \ref{sec:engineering}, we propose that the software development lifecycle provides an appropriate model for the dataset development lifecycle. This builds on previous analyses of the machine learning lifecycle (\eg{} \cite{zaharia2018, ashmore2019}) and the data science lifeycle \cite{boehm2019}, by exploding what is sometimes considered a mere single stage of ``Data Management''.\footnote{We differ from the ``data lifecycle'' analysis of \cite{polyzotis2018} in that they consider the ``first step of ML is to understand your data'', presuming that data already exists.}

\renewcommand{\arraystretch}{1.5}

\subsection{Lessons from engineering models}
Alvi proposes a set of recommendations for engineering models (Section~\ref{sec:engineering}) \cite{alvi2013engineers}, and they are equally applicable to datasets. We summarize them here, substituting ``dataset'' for ``model'' for clarity.
\begin{itemize}
\item Treat datasets as \textit{guilty until proven innocent}.
\item Identify the assumptions underlying a dataset in writing.
\item Conduct independent peer reviews and checks during and after dataset development.
\item Evaluate datasets against experience and judgment.
\item Use visualization tools.
\item Perform sensitivity studies of dataset parameters. 
\item Understand the assumptions and limitations of datasets, rather than using them as black boxes.
\end{itemize}
The model of dataset development  documentation that we present in the following section is aligned with these recommendations.

\subsection{Documentation of Dataset Development}
\label{sec:documentation}

Documentation of the stages of the dataset development process serve as \textit{accounts}, in the sense just described above. 
As such, we now delve into the documentation artefacts relevant to each of the lifecycle stages (Table~\ref{tab:documents}). Although the precise number and nature of lifecycle stages is sometimes debated, for the current purposes it is sufficient to consider the  5 stages of the lifeycle in Table~\ref{tab:stages}. Each document has owners with designated roles and duties \cite{krafft2020}. 

\begin{table}
\centering
\begin{tabular}{p{6em}|p{18em}}\hline
{\it Requirements analysis} & Deliberations about intentions, consultations with stakeholders, and analysis of use cases determine what data is required. \\

{\it Design} & Research is performed and subject matter experts are consulted in order to determine whether the data requirements can be met, and if so how best to do so. \\

{\it Implementation} & Design decisions are transformed into technologies such as software systems, annotator guidelines, and labeling platforms. Actions may employing and managing teams of human expert raters. \\

{\it Testing} & Data is evaluated  and decisions about whether or not to use it are made.\\

{\it Maintenance}& Once collected, a dataset requires a large set of affordances, including tools, policies and designated owners.\\\hline
\end{tabular}
\caption{Stages of the Data Development Lifecycle.}\label{tab:stages}
\end{table}

\subsubsection{Data requirements accounts} 

Why is data the solution to a problem at hand? Given that data is infrastructure (see Section~\ref{sec:infrastructure}), and that datasets are engineering artefacts (Section~\ref{sec:engineering}), it is important to consider the ends to which data is the means, and (to avoid data-solutionism) whether data is the best way of addressing this problem. Given the long history of existing data being put to novel uses, one must also account for the consequences of  unintended uses. The thoroughness of such accounts speaks directly to questions of \textit{ex ante} accountability. The 6 W's provide a useful framework for detailing such intended and unintended scenarios, \ie{} specifying for each use the who (the decision-makers) and the whom (those being impacted), as well as the what, where, when and why \cite{wieringa2020}.

In engineering, these considerations all fit under requirements analysis \cite{barclay2019towards}, and we embrace the proposals of \cite{loshin2010master,loshin2010practitioner, vogelsang2019requirements} for \textbf{Data Requirements Specifications}---directly analogous to software requirements---covering both quantitative and qualitative factors. One of the key tasks is holding stakeholder requirements development sessions in which the (possibly conflicting) needs of multiple stakeholders are collected and aggregated, and conflicts resolved through accountable mechanisms (including the keeping of accounts regarding the conflicts), making transparent what is valued in the data \cite{walmsley2020artificial}.
As with software requirements, a number of challenges may exist when documenting dataset requirements. Stakeholders may not communicate their needs, or may be reluctant to commit to a set of written data requirements. Data development teams may have a natural inclination to ``jump in'' and start collecting data before the requirements analysis is complete.  Stakeholder groups may have different vocabularies, or may be unable to reach consensus using previously agreed decision-making mechanisms. Stakeholders may try to make the requirements fit an existing dataset, rather than think through their specific needs. Cognitive biases and standpoints may prevent complete accounts.

The task of producing requirements documents should have a clearly designated owner.  The frequent neglect of clearly specified requirements for data goes beyond ML data  (\eg{} \cite{cmmi2018}). There is likely a need to encourage and reward the development and ownership of requirements specifications, by making this part of a key task, rather than a new burden placed on top of already demanding schedules \cite{sambasivan2020}.\footnote{This burden is common to many fields: ``faced with ever more crowded diaries, we increasingly subsidize data collection to a very significant degree through our personal time and often through enormous personal sacrifice'' \cite{lane2020editorial}.} Since the benefits and risks of infrastructure accrue over time, the long horizons of the impacts of data requirements analysis need to be taken into account when judging success or failure. In other words, both the quality and quantity of uses of the dataset, over long time horizons, need to be taken into account.

Appendix A provides a flexible template for documenting dataset requirements, adaptable to many contexts and organizational environments using processes of contextual inquiry \cite{bednar2009contextual}. It covers a broad range of requirements about data instances, distributions, processing and sourcing, motivation and intended uses, as well as critical metadata such as approvers and changelog.

\subsubsection{Dataset design accounts} 
\label{sec:documentation_design}

The distinct roles of requirements and design are sometimes misunderstood,\footnote{And as \cite{kuwajima2020engineering} drily observe, machine learning is characterized by a lack of both requirements specifications and design specifications.} but it is important to emphasise to stakeholders the conceptual distinction between requirements and design, to avoid premature decisions about design before requirements are established. Whereas requirements analysis sets the goalposts for the dataset, the design answers the question of means: \textit{how will we get there}? That is, the former answers \textit{what?} and \textit{why?} by documenting ends and intentions, while the latter answers \textit{how?} by documenting means. To give a software analogy, while a requirement may be that a function return a sorted list, a design decision may be to use the bubble sort algorithm. Research of existing options is a critical part of the design phase. If a pre-existing dataset (or combination of datasets) is satisfactory, there is usually no good reason to design and build a new one. This stage may therefore necessarily involve elements of Dataset Testing (see below), as pre-existing datasets are evaluated against the current requirements.

The primary account of this stage is the \textbf{Dataset Design Document}. This document's primary roles are to lay out the plan of how requirements will be achieved, and to justify the design decisions that are made. These justifications take myriad forms, but consultation of domain experts is a critical part of this stage. Crystal clear and objective options are often lacking, but rather the formation of information into discrete cases involves decisions about standardization and lossiness \cite{busch2014big}.  Since the future lives of datasets are impossible to predict (and ``researchers in neighboring fields may view the same data in considerably different ways'' \cite{busch2014big}, see also \cite{pasquetto2017reuse}), creating detailed accounts of assumptions and decisions is critical to the dataset design phase. Documenting tradeoffs and which alternatives were considered---both technical and nontechnical---are central to the design process, and integral to engineering practice \cite{goldman2010beyond}. Even seeming ``neutral decisions'', such as how to sample, require multiple decisions: datasets require formation of cases, which require categories, and these categories have consequences that are frequently statistical and sometimes moral. Aspects of phenomena which are easily calculated become amplified, while those which resist standardization are reduced or excluded cited by \cite{busch2014big}. 

The field of Natural Language Processing's historical use of carefully constructed balanced datasets---prior to the proliferation of internet data \cite{kilgarriff2003introduction}---points to lessons about dataset design which remain relevant. These include the necessity of detailed discussions of ``what it means to `represent' a [phenomenon]'', ``definition of the target population,''
and consideration that ``theoretical research should be prior'' in dataset design ``to identify the situational parameters,'' after which construction of the dataset ``would then proceed in cycles''
\cite{biber1993representativeness}.

Appendix B provides a flexible template for documenting dataset design decisions, adaptable to many contexts and organizational environments. It includes sections detailing related datasets, how data will be sourced and annotated, key empirical characteristics, how privacy will be handled, as well as how data quality will be measured.

\subsubsection{Dataset implementation accounts} 
Regardless of the combination of human labour and computation involved in dataset creation, more decisions have to be made---and documented---during dataset implementation. To extend the previous comparison, if the design phase specifies the bubble sort algorithm is to be used, the implementation might choose data structures and control flows whose appropriateness may be contextually dependent. An analogy of dataset implementation accounts is therefore with code comments: fine-grained documentations tightly coupled with the implementation. Since digital traces can often be mined for the answers to \textit{what was done}, as with code comments the most important implementation accounts are for explaining \textit{why things were done that way} \cite{sutter2004c++}. There are many ways that these accounts of data collection and labeling can be recorded. As with code comments, tight coupling has benefits of synchronicity; that is, tools that are used to manage and support data collection and labeling should ideally have integrated commenting capabilities, with comments being exportable and searchable. Failing such tight integration, a temporally organized \textbf{Dataset Implementation Diary}, can be used. Issue-tracking systems can also be adapted for this role, e.g. with each decision being blocked until justified on the appropriate ``ticket''.

\subsubsection{Dataset testing accounts} 
Data needs testing just like code does \cite{vogelsang2019requirements}.
As with software testing, there are many different flavours of dataset testing, but at their foundation is the question \textit{should this data be used}? A \textbf{Dataset Testing Report} is the account that specifies the evaluations that were done, as well as their results. These are naturally  analogous to software unit testing and security testing. In both cases, accounting of the diligence and coverage of the test cases provides the ``receipts'' for justifying trust in the data. At the lower levels of testing are questions of data wellformedness (for which a range of tools are available \cite{ehrlinger2019survey}); at higher levels, questions of fidelity \cite{beretta2018ethical}. Two particularly important types of testing are requirements testing and adversarial testing.  

\textit{Requirements testing} (also called ``acceptance testing'' \cite{glinz2011glossary}) checks a dataset  against the stated requirements for its use. (These can be the requirements which led to the creation of the dataset, but alternatively a pre-existing dataset may be evaluated against a novel set of requirements.) Requirements testing directly reframes the question  of ``how to evaluate whether the right data was collected with sufficient quantity'' \cite{roh2019survey}, by checking whether we got what was needed. As with software testing, ideal data testing returns boolean pass/fail outcomes; difficulties in doing so may point to a lack of clarity in the dataset requirements. However it is not necessary that all inputs to testing be present in the data records themselves; evidence can also come from metadata generated throughout the data collection lifecycle. For example, an intention to share a new dataset may lead to certain requirements in how consent is obtained, and the data collection process needs to keep ``the receipts'' as evidence of due process. To facilitate this, it is therefore necessary that dataset requirements be \textit{traceable} \cite{mittelstadt2016ethics}, and in many  cases requirements should be measurable. Non-boolean test outcomes are still useful in flagging concerns, and frequently this can lead to discussion and iteration on requirements. When requirements and their testing is specified clearly enough, opportunities for test automation become possible, including realtime monitoring and alerting \cite{breck2017ml}. When tests have sufficient validity and coverage, automated testing can provide reliable guarantees for continuous data deployments.  Automated tests can also feed back into the data collection stage, providing signals as to which parts of the data distribution need further sampling. When such tests involve ML model evaluation, this takes on the flavor of active learning \cite{settles2009active}. 

\textit{Adversarial testing} of a dataset (also known as ``test-to-fail'' testing \cite{patton2006software}) aims to uncover unforeseen harms which may come from its use. Just as in software  security, the tester plays the role of an ``attacker'' and tries to break things by finding risks of undesirable outcomes. Such risks vary widely in form, from harms to specific individuals such as privacy leaks, to harms to subgroups such as unforeseen correlations or stereotypes, to public relations risks via embarrassing data (including errors or omissions), to the possibility of malicious third parties using the data for nefarious ends. 

Appendix C contains a template for Dataset Testing Reports, to summarize the processes and results of testing.

\subsubsection{Dataset maintenance accounts} \label{sec:maintenance}
If you can't afford to maintain a dataset, and you also can't afford the risks of not maintaining it, then you can't afford to create it.
However a common challenge with many forms of infrastructure is that funding for creation is often more readily available than for maintenance, leading to a maintenance budget gap. 
Furthermore, charitable volunteerism cannot be expected as a substitute, since maintenance is one of the least desirable roles in the software lifecycle \cite{capretz2015influence} and the same is likely true of data.
Indeed since many maintenance tasks result in no visible impacts beyond preservation of the \textit{status quo},  maintenance work risks being viewed as an unrewarded, unglamorous burden. 

A forward-looking \textbf{Dataset Maintenance Plan} is the appropriate account for this stage (analogous to a software maintenance plan \cite{abran2004software}). Although adding new requirements is sometimes called ``perfective maintenance'' \cite{bennett2000software}, we consider that task to be a fresh iteration of the requirements phase of the lifecycle; the maintenance plan should instead focus on corrective, adaptive and preventive  goals \cite{bennett2000software, ieee1990ieee}. \textit{Corrective maintenance} aims at fixing errors. Unforeseen problems are to be expected, and affordances for human-data interaction should thus be provided. Data poisoning may have occurred, labeling guidelines may have been misconstrued, or the phenomena in question may quite simply change. Affordances for manual edits and batch updates need to be in place (with appropriate access controls), and all such actions need to be scrupulously logged. Making these logs human-readable and searchable is an important step in dataset maintenance accountability. \textit{Adaptive maintenance} aims at preserving the key properties of a dataset under changing external conditions. The critical question is thus: \textit{what properties should be preserved?} Here requirements come into play: one key goal is preserving the dataset's satisfaction of its own requirements. Challenges arise when drift occurs, \ie{} the dataset diverges from the reality of the phenomena it purports to measure. Addressing this requires feedback loops with other lifecycle stages, to set up recurrent measurement of the validity and fidelity of the data. \textit{Preventive maintenance} aims to prevent problems before they occur, for example paying down technical debt before it becomes unwieldy (see Section~\ref{sec:discussion}).

Datasets should be stored or maintained in stable, institutional repositories that allow for differential levels of access and stable universal identifiers. While this may be less of an issue in organizations with mature internal data infrastructures (such as corporate firms), this can present a special problem with datasets which are ostensibly released for use for the research community, and yet are hosted on personal and lab websites. In the research space, stable institutional repositories used for social science research include the Harvard's Dataverse \cite{king2007introduction}, University of Michigan's ICPSR \cite{eulau2007crossroads}, and New York University's Databrary \cite{gilmore2016curating}. These are robust projects with years of institutional support, which decreases the infrastructural cost for hosting by lower-resource organizations and teams.

Maintenance costs need to be continually weighed against benefits, and eventually there may come a time when a dataset should no longer be maintained but instead deleted. The easy replicability of binary data means that deleting all copies of a dataset becomes impossible once it has been shared. So in the absence of complex cryptographic solutions, dataset sharing should  be tightly controlled whenever contestability is required (see Section~\ref{sec:welfares}).

\subsection{Dataset Audits and Reviews}
\label{sec:audits}
As discussed above, accountability for datasets to the forum requires discussion and deliberation of the dataset documentation (i.e. ``accounts''). One type of deliberation process is the audit, which in ML contexts might review datasets and/or models. In internal ML audits, initial scoping and mapping phases are followed by an artefact collection phase, prior to evaluation and reflection by the audit team during which the artefacts are analysed in light of the organization's policies and commitments \cite{raji2020closing}. The dataset development documentation described above form part of the audit trail \cite{us2018data}, boosting auditability by design \cite{zook2017ten}. The audit is but one form of review, however, and just as for software the entire dataset lifecycle requires reviewing (\eg{} requirements reviews, design reviews, \etc{}), especially during initial development, but also when contextual circumstances change.

\subsection{Lessons from Infrastructure Governance}
Governance mechanisms for infrastructure can inform accountability mechanisms for datasets. We focus here on public infrastructure governance, but models of private dataset governance also exist \cite{loshin2010practitioner}. Like dataset projects, public infrastructure projects have often begun with poor understanding of risks and challenges, and with an underappreciation of the work required. This spirit of entrepeneuralism risks dangerous adventurism \cite{anheier2017infrastructure}, and misestimations of both costs and benefits \cite{flyvbjerg2016principle}. To mitigate these, risk anticipation and management should be prioritized \cite{anheier2017infrastructure}.
One governance mechanism is the stakeholder consultation process in dataset requirememts analysis, however a realistic understanding is required of how contested decisions are resoved \cite{wegrich2017infrastructure}. One tradeoff is between hierarchical coordination (appeal to institutional authority) and negotiated coordination (\eg{} by requiring consensus). The former often involves appeals to some form of welfare-maximization, while the latter may restrict the decision space to Pareto optimal solutions. 
It may take years for problems in datasets  to be discovered
(\eg{} Imagenet was developed in the late 2000s, with certain  issues not identified until 2019 \cite{crawford2019excavating}). This supports calls for social accountability mechanisms which emphasise legitimacy via inclusivity in analysis and decision-making phases \cite{jordana2017accountability}, including focus groups, town meetings, public hearings, and randomised citizen involvement.
Identification of ``territories'' of interests enables conversations about how risks and benefits are accrued to various groups. Temporal discounting and inter-generational effects may also be relevant: are there future effects of datasets on those who are too young to have a say today?

\section{Discussion}
\label{sec:discussion}

In Section~\ref{sec:documentation} we presented a model of dataset development documentation which facilitates transparency, motivated by accountability's dependency on visibility. We now discuss the broader systems of dataset development itself, and the cultural systems in which dataset work is embedded.

\subsection{Benefits of Data Documentation}
\label{sec:welfares}

Just as with software development \cite{zhi2015cost}, there are diverse costs and benefits as to how dataset development documentation interacts with project comprehension, management decision-making, and time efforts/savings. More empirical evaluations of these are needed for data documentation. Improved documentation helps developers with all stages of the lifecycle but, just as with software documentation \cite{zhi2015cost}, we expect that improved documentation will particularly aid dataset developers in the maintenance phase, due to both collective amnesias and employee turnover.

\paragraph{Data Work Recognition}
Documentation increase employees' valuation of dataset work by ``outlining the nature of an individual or team’s contribution to an overall system'', creating  space for reflection, debate and recognition, and by giving ``meaningful understanding of the impact of their personal participation''  \cite{raji2019ml}. Open data advocates have argued that data availability has led to increased data citation, which can serve as a professional incentive for doing necessary data work, as well as developing data documentation \cite{piwowar2013data}.

\paragraph{Dataset Productionization}
Productionized deployments of ML models require much more than evaluating model performance metrics in isolation \cite{breck2017ml}. In these contexts, the model itself will form only one part of the deployed system, requiring infrastructure for deployment and monitoring \cite{baylor2017tfx}. Dataset development documentation outlined above plays a critical role in defining the contracts between different components, helping product developers, maintainers and end-users determine how much trust to place in the system \cite{arnold2019factsheets}. Additionally, in online ``continuous training, continuous serving" systems, the concept of a ``dataset" as a static artifact (or even a versioned artefact) breaks down. Instead, datasets are defined by automated pipelines of data processing steps, requiring automated testing  \cite{sculley2015hidden} against requirements and expectations, along with monitoring and alerting.

\paragraph{Data Debt Mitigation}
The deferred engineering maintenance cost that occurs when speed of execution is prioritized over quality of execution is known as ``technical debt''. Like financial debt, technical debt compounds, ultimately slowing down long term progress in pursuit of short term gains (sometimes leading to the ``Winner's Curse'' \cite{sculley2018winner}). Technical debt is equally relevant to datasets, in fact dataset dependency debts can be more costly to maintain than code dependencies \cite{sculley2015hidden}. Critical to addressing this are maintenance practices that include budgeting for debt repayment, as well as explicit definitions and measurements of quality, via data requirements documents and data testing reports. Undesirable data feedback loops and unutilized data features can be mitigated with careful requirements practices.  
\paragraph{Data Postmortems and Premortems}
Even the best designed systems occasionally fail \cite{rittel1973dilemmas, buchanan1992wicked}, however cyclical development practices help avoid making the same failures twice, and the postmortem is a critical tool in learning from past failure and building institutional memory \cite{dingsoyr2005postmortem}. Institutional debugging  of ``what went wrong'' with datasets requires both a shared definition of what ``right'' looks like (\ie{}  requirements), as well as rigorous testing and test validation (for faulty tests are worse than no tests \cite{osherove2015art}).
A premortem involves imagining in advance what could go wrong \cite{klein2007performing}, and, similarly for datasets, requires a shared definition of ``right''. Indeed, premortems can be a useful tool during data requirements analysis for identifying gaps and vagueness.
\paragraph{Transparent Dataset Reporting}
Transparent reporting of machine learning datasets \cite{gebru2018datasheets, bender2018data, holland2018dataset} and the models trained on those data \cite{mitchell2019model} enables third party agents, including potential users, to make informed decisions. Poor reporting practices result in organizations not having the tools they need to understand ML datasets prior to building models \cite{holland2018dataset}. 
Careful documentation throughout the lifecycle creates the intra-institutional memory required for transparent reporting to third parties.

\paragraph{Data Reuse and Reproducibility}
Open reporting of machine learning datasets allows third parties, such as auditors and researchers, to attempt replication of results and predictions, thereby strengthening trust in quantitative findings. This work coalesces well with the movement within computational sciences who have called for standards to include data and code with research findings, oriented towards the dissemination of reproducible research, facilitating innovation, and enabling broader communication of scientific outputs \cite{stodden2014best}.

\paragraph{Data Contestability}
Designing mechanisms of contestation into algorithmic decision making systems has been motivated as a way of encouraging critical and responsible engagement between various stakeholders, surfacing values that might not otherwise be visible, and safeguarding a system against misuse \cite{mulligan2019shaping, hirsch2017designing}. However, the ability of a given stakeholder to challenge an automated decision is limited, in large part, by the degree to which different components of the algorithmic system are legible to those outside the development team. By surfacing otherwise invisible aspects of dataset development the proposed data documentation frameworks creates the conditions necessary for contesting datasets as well as algorithmic systems built from them.  

\paragraph{From Metadata to Data}
Information about processes of dataset development constitutes metadata, \ie{} structured data about data \cite{us2018data}. This metadata can in some cases be incorporated \textit{into the dataset itself}, analogous to how digital image files record time, date and camera settings \cite{camera2010exchangeable}. In the case of data labeling, this collapses the dichotomy of the the observer and the observed, making label provenance  explicit in the data \cite{stasaski2020more} rather than pretending to be a ``view from nowhere'' \cite{nagel1989view, haraway1988situated} and creating the space for alternate epistemologies needed to confront the bias paradox \cite{heikes2004bias}.  Reconceiving dataset records as essentially relational artefacts, between the beholder and beheld, enables datasets to document aspects of their own creation. Making epistemological uncertainties in data explicit in the data itself somewhat parallels engineering cultures of exception-handling, in which classes of errors are thrown and caught (see Section~\ref{sec:engineering}), \ie{} it is the \textit{unknown} errors which are the most dangerous in safety-critical applications.\footnote{The Apollo missions of the 1960s “flew with program errors that were thoroughly documented, and the effects were understood before the mission. No \textit{unknown} programming errors were discovered during a mission.” \cite{hall1996journey} (original emphasis).}

\subsection{Ecologies of Data Work}
Based on our analysis above of the processes needed to improve dataset development documentation, there are a number of ecological challenges that potentially inhibit process adoption. To mitigate these challenges, we propose the following goals for research, academia and industry. 

\paragraph{Recognize AI dataset expertise}
AI datasets take more effort and rigor to curate and manage than code does \cite{amershi2019software}, and yet this work remains underappreciated.
Omitting dataset development from the core training of ML practitioners is comparable to not teaching computer programmers about data structures.
Following Jo and Gebru, we call for cultivating \textit{AI Dataset Development} as a distinct subfield of AI (with parallels to archival studies and corpus linguistics) \cite{jo2020lessons}, providing scaffolding that serves as a primary and fundamental input to the development lifecycle. Seen through this lens, dataset development requires a great deal of further dedicated research (\eg{} on practices of collection, methods and protocols), its own theory  \cite{coveney2016big},\footnote{Especially since academia values theory over practice \cite{goldman2004we, hacking1983representing}.} conferences, prizes, \etc{}
There is a need to shift the perception of dataset work away from crafty gluework to highly skilled technical infrastructure work. Data provides a fundamental scaffolding from which further ML development can launch.  Recasting AI dataset work from drudgework or necessary evil to valued careful cultivation and stewardship means acknowledging the roots of dataset work in feminized practice, notably informed by proximate fields in library and information studies \cite{hoffmann2016digitizing}.
Some seeds worth cultivating include values-driven active learning \cite{anahideh2020fair},  diversity-informed data collection \cite{stasaski2020more}, and challenge data sets \cite{davis2014limitations}.\footnote{These currently seem most common in NLP, \eg{} \cite{khashabi2018looking,mullenbach2019nuclear, yang-etal-2015-wikiqa, richardson-etal-2013-mctest, liu2020logiqa, williams-etal-2018-broad, yagcioglu2018recipeqa, zellers2018swag}.}

\paragraph{Engage with data disciplines beyond ML}
Building on our call for engagement with subject domain experts when building datasets (Section~\ref{sec:documentation_design}), we also call for engagement with disciplines with rich dataset development practices. We echo calls to learn from archival studies
\cite{jo2020lessons} and language corpus design \cite{xiao2010corpus}, and to pay greater attention to research in human computation \cite{vaughan2018}. 
AI Dataset Development could also benefit from engagement with established data quality and data integrity practices from clinical research. In medicine data integrity refers to the completeness, consistency, and accuracy of data, and to achieve these data should be attributable, legible, contemporaneously recorded, original or a true copy, and accurate ({\sc ALCOA}) \cite{food2016data}. The US FDA states that ``System design and controls should enable easy detection of errors, omissions, and aberrant results throughout the data’s life cycle'' \cite{us2018data}. What could analogous practices look like for AI dataset development? 

\paragraph{Understand cultures of dataset development}
There is need for more research on the cultures of AI Dataset Development, echoing work on cultures of computer programming (see Section~\ref{sec:engineering}) and cultures of mathematics \cite{livingston1999cultures}. Recent work in this direction includes Denton and Hanna \etal, who discuss focusing on ``bringing the people back in" to the discussion of dataset construction \cite{denton2020bringing}. Better understanding is needed of how the deeply collaborative nature of much dataset work conflicts with ``disincentives to collect data that come from a system that emphasizes individual academic output'' \cite{lane2020editorial}, as well as relationships with other common ML myopias---especially common to the ``hacker culture'' of ML (Section~\ref{sec:engineering})---that celebrate Alpha leadership, competition, and other masculine tropes.\footnote{The connection of "engineering" itself with computer programming was an intentionally, highly gendered endeavor intended to shift work from a primarily feminized craft to something that would pay more and attract men \cite{abbate2012recoding}.}

\paragraph{Invest in organizational processes and structures}
Adopt best practices from managerial theory to software development, including critically the needs for properly budgeting for maintenance and treating it properly as ``repair work'' \cite{sachs2019algorithm},  for redressing dataset technical debt, and for organizational recognition and rewards for data work. We need to shift processes such that explicit normative considerations are no longer rarely in mind \cite{passi2019problem}.

\paragraph{Demand greater methodological clarity of ML research}
Whereas methods are tools of research, \textit{methodologies} are commonly described as the principles, values or frames of reference of research \cite{mackenzie2006research}.  
The methodologies of ML research are often opaque or obscure, \eg{} is the goal to acquire \textit{knowledge-how} or \textit{knowledge-that} (Section~\ref{sec:engineering})? Is the research akin to mathematics, science, applied science, engineering,  engineering science, or something else? These questions matter not just because they bring justification to methods (why are robustness, AUC, effect sizes or $p$-values the relevant things to measure?), but also because the ML goals determine the ML dataset requirements. ML datasets need to be understood as engineered artefacts in support of ML goals; if we instead mistake dataset development for ``academic science", we perpetuate dangerous assumptions about objectivity.

\section{Conclusions}
We have described the role of data in AI system development and highlighted the gap between current development processes and processes that can address modern AI outcomes. We ground development within concrete ethical practices, proposing frameworks to operationalize transparency and accountability. Successful implementation of our proposed frameworks depends critically on clearly formulated development processes, with discrete and interconnected stages. Owners charged with responsibility for each stage must work by consulting appropriate experts and produce detailed accounts of what happens during each stage, and why. The form of these accounts (requirements specifications, design documents, testing reports, \etc{}) actively work to improve the quality, reliability and validity of the data itself.

Our approach is informed by  the software development lifecycle, based on two correspondences that we detailed above: 1) between data and technical infrastructure, and 2) between the processes and goals of engineering with those of dataset development. Based on these similarities, we presented a clear and comprehensive framework that builds on existing best practices. We leverage methods in related fields to describe a procedure for documenting all stages of dataset development: requirements analysis, design, implementation, evaluation, and maintenance. The detailed bookkeeping of accounts of each stage of the lifecycle requires owners with distinct knowledge, skills, and processes, and the expertise most critical to consult at each stage is similarly distinct: stakeholders and impacted parties; subject matter experts; HCI, data operations managers and raters with domain expertise; data scientists; supporters of tooling and ongoing funding.

We argue that frameworks for transparency and accountability are badly needed, as systemic devaluation of dataset work has contributed greatly to its historic failures. We believe that a culture shift is needed to truly embrace the modernization of AI development processes towards ethically-informed frameworks, We advocate for valuing work on data commensurate with the value placed on technical infrastructure work and model building. If we are to improve accountability for the datasets that power ML, then, perhaps paradoxically, we need to create avenues for which datasets can earn accolades when they are properly due, avenues which encourage skilful data experts to proudly say ``I am responsible for this.''

\bibliographystyle{ACM-Reference-Format}
\balance
\bibliography{references}

\begin{table*}
\caption*{\Large APPENDIX A: TEMPLATE FOR DATASET REQUIREMENTS SPECIFICATION }

\begin{tabularx}{\textwidth}{|XX|}
\hline
&\\\multicolumn{2}{|c|}{\templatetitle{{\em Name of Dataset}: Requirements Specification} }\\
&\\\multicolumn{2}{|c|}{Owner: \textit{Name}; Created: \textit{Date}; Last updated: \textit{Date}}\\

\begin{tabular}[t]{p{7.5cm}}
\templatesection{Vision}
Brief summary of the envisioned data(set), its domains and scope.

\templatesection{Motivation}
Problem and context that motivate why the data is needed.

\templatesection{Intended uses}
Specific uses of the data that are intended.

\templatesection{Non-intended uses}
What is the data not intended for? What should the data not be used for, and why?

\templatesection{Glossary of terms}
If relevant, brief summary of acronyms and domain specific concepts for the general reader. 

\templatesection{Related documents}
List any related documents.

\templatesection{Data mocks}
Include 2-3 typical examples of what the data instances should "look" like.

\templatesection{Stakeholders consulted}
Whose needs were consulted and synthesised when creating this document? How were conflicting needs resolved?

\templatesection{Creation requirements}
Where should the data come from? Include sources and collection methods

\begin{itemize}
    \item {\em Name of the requirement. Description.}
    \item {\em Name of the requirement. Description.}
\end{itemize}

\templatesection{Instance requirements}
What requirements are there for data instances? Include any acceptable tradeoffs. Include numbers and types of instances, features, and labels.

\begin{itemize}
    \item {\em Name of the requirement. Description.}
    \item {\em Name of the requirement. Description.}
\end{itemize}

\\
\end{tabular}
&
\begin{tabular}[t]{p{7.5cm}}
\templatesection{Distributional requirements}
What requirements are there for the distributions of your data? Include any acceptable tradeoffs. Include sampling requirements. If your data represents a set of people, describe who should be represented and in what numbers. 

\begin{itemize}
    \item {\em Name of the requirement. Description.}
    \item {\em Name of the requirement. Description.}
\end{itemize}

\templatesection{Data processing requirements}
How should the data be annotated and filtered? Who should do the annotating? How should data be validated? Include any acceptable tradeoffs.

\begin{itemize}
    \item {\em Name of the requirement. Description.}
    \item {\em Name of the requirement. Description.}
\end{itemize}

\templatesection{Performance requirements}
What can people who use this dataset for its intended uses expect? 

\begin{itemize}
    \item {\em Name of the requirement. Description.}
    \item {\em Name of the requirement. Description.}
\end{itemize}

\templatesection{Maintenance requirements}
Should the data be regularly updated? If so, how often? For how long should the data be retained? Include any acceptable tradeoffs.

\templatesection{Sharing requirements}
Should the data be made available to other teams within Google and/or open-sourced? If so, what constraints on data licensing, access, usage, and distribution are needed? Include any acceptable tradeoffs.

\templatesection{Caveats and risks}
What would be the consequences of using data meeting the requirements described above?

\templatesection{Data ethics}
Document your considerations of the ethical implications of the data and its collection.

\\
\end{tabular}
\\
\begin{tabular}{|p{0.13\textwidth}|p{0.13\textwidth}|p{0.13\textwidth}|}
\multicolumn{3}{c}{\Large Sign-off grid} \\\hline
\rowcolor{Gray}
Name & Role & Date\\\hline
\rowcolor{white}
&&\\\hline
\end{tabular}
&
\begin{tabular}{|p{0.13\textwidth}|p{0.13\textwidth}|p{0.13\textwidth}|}
\multicolumn{3}{c}{\Large Changelog} \\\hline
\rowcolor{Gray}
Editor & Comments & Date\\\hline
\rowcolor{white}
&&\\\hline
\end{tabular}
\\
&\\\hline
\end{tabularx}
\end{table*}

\begin{table*}
\caption*{
\begin{centering}
{\Large APPENDIX B: TEMPLATE FOR DATASET DESIGN DOCUMENT }
\end{centering}
}
\textmd{When adapting the template, we recommend reflecting on questions such as ``\textit{If you were asked to review the design of this dataset, what questions would you ask?}” and ``\textit{If you were presenting a design, what questions would you dread being asked?}” \cite{hall2014distributed}.}
\\
\vspace{6pt}
\begin{tabularx}{\textwidth}{|XX|}
\hline
&\\\multicolumn{2}{|c|}{\templatetitle{{\em Name of Dataset}: Design Document} }\\
&\\\multicolumn{2}{|c|}{Owner: \textit{Name}; Created: \textit{Date}; Last updated: \textit{Date}}\\
\begin{tabular}[t]{p{7cm}}
\templatesection{Overview}
\templatetext{High-level overview of the dataset.}

\templatesubsection{Dataset Name}
\templatetext{Name of dataset.}

\templatesubsection{Primary Data Type(s)}
\templatetext{Primary data types; Eg: images, video, text.}

\templatesubsection{Data Content}
\templatetext{Eg. bounding boxes, image labels}

\templatesubsection{Funding}
\templatetext{How was the dataset funded?}

\templatesection{Objective}
\templatetext{What are the key objectives of the dataset? Is there a requirements specification?}

\templatesection{Version}
\templatetext{Current version; Differences to previous versions.}

\templatesection{Background}
\templatetext{Describe any relevant background of the dataset}

\templatesection{Sources}
\templatetext{System details; Where will the data come from?  Selection and sampling criteria.}

\templatesection{Annotations}
\templatetext{Features and labels; Who are the annotators? How will they be trained?}

\templatesubsection{Ratings}
\templatetext{Rating tasks; Rating types; Rating procedures;}

\templatesection{Data Quality}
\templatetext{How is quality measured? How are metrics validated?}
\end{tabular}
&
\begin{tabular}[t]{p{7cm}}
\templatesection{Characteristics}
\templatetext{Characteristics of the dataset.}

\templatesubsection{Expected Characteristics}
\templatetext{Eg. How many instances, features, ratings.}

\templatesubsection{Correlations}
\templatetext{Acceptable correlations; Unacceptable correlations;}

\templatesubsection{Acceptable and Unacceptable Conjunctional Datasets}
\templatetext{Datasets that can and can not be used in conjunction with this dataset?}

\templatesubsection{Population}
\templatetext{Population represented.}

\templatesection{Privacy Handling}
\templatetext{How is privacy handled?}

\templatesection{Maintenance}
\templatetext{Who will maintain the dataset? Is there a maintenance plan? What are the recovery strategies if issues arise?}

\templatesection{Sharing}
\templatetext{Will the dataset be shared? How will access be controlled? How will the dataset be licensed?}

\templatesection{Caveats}
\templatetext{Describe known caveats}

\templatesection{Data Ethics}
\templatetext{Ethical considerations; Mitigation.}

\templatesection{Work estimates}
\templatetext{How much time will it take to collect the data; Costs are involved.}
\end{tabular}
\\
\multicolumn{1}{|X}{
\begin{tabularx}{\textwidth}{X}
\templatesection{Related Datasets}
\templatetext{Which existing datasests are related to this one? Why are they unsuitable? }

\templatesubsection{Dataset Discovery Process}
\templatetext{How did you search for other datasets?}

\templatesubsection{Survey}
\templatetext{High level overview of related datasets.}

\begin{tabular}{|m{4cm}|m{3.5cm}|m{3.5cm}|m{3.5cm}|}
      \hline
      \rowcolor{Gray}
      & \textbf{\small{Your  Dataset}}
      & \textbf{\small{Other Dataset 1}}
      & \textbf{\small{Other Dataset 2}} \\ \hline
      \rowcolor{white}
      \textbf{\small{Documentation and DOI}}
      & \small{Yours}
      & \small{Theirs}
      & \small{Theirs} \\ \hline
      \textbf{\small{Motivation and Intended use}}
      & \small{Yours}
      & \small{Theirs}
      & \small{Theirs} \\ \hline
      \textbf{\small{Location}}
      & \small{Yours}
      & \small{Theirs}
      & \small{Theirs} \\ \hline
      \textbf{\small{Size, Sampling and Filtering}}
      & \small{Yours}
      & \small{Theirs}
      & \small{Theirs} \\ \hline
      \textbf{\small{Annotation and Labels}}
      & \small{Yours}
      & \small{Theirs}
      & \small{Theirs} \\ \hline
      \textbf{\small{(Expected) Performance}}
      & \small{How will your dataset improve on existing ones?}
      & \small{How does it not meet your requirements?}
      & \small{How does it not meet your requirements?} \\ \hline
      \textbf{\small{Examples}}
      & \small{Example instance}
      & \small{Example instance}
      & \small{Example instance} \\
      \hline
\end{tabular}

\end{tabularx}
}
&\\
\hline
\end{tabularx}
\end{table*}

\begin{table*}
\caption*{\Large APPENDIX C: TEMPLATE FOR DATASET TESTING REPORT }

\begin{tabularx}{\textwidth}{|X|}
\hline
\\
\multicolumn{1}{|c|}{\huge \textit{Name of Dataset}: Testing Report}\\
\\ \multicolumn{1}{|c|}{Owner: \textit{Name}; Created: \textit{Date}; Last updated: \textit{Date}}\\

\templatesection{Summary}
\templatetext{What is being tested?}

\templatetext{Link to requirements specification.}

\templatetext{Link to design document.}

\templatesection{Meta-Testing}
\templatetext{Is the data still needed? Are the data requirements still relevant and up-to-date?}

\templatesection{Requirements Testing}
\begin{tabular}{|m{4cm}|m{4cm}|m{8cm}|}
      \hline
      \rowcolor{Gray}
      \textbf{\small{Requirement tested}}
      & \textbf{\small{Results}}
      & \textbf{\small{Artifact}} \\ \hline
      \rowcolor{white}
      \small{Requirement from requirements specification}
      & \small{Score or Results}
      & \small{Justification of the results or a link to artifact}
      \\ \hline
      \small{Requirement from requirements specification}
      & \small{Score or Results}
      & \small{Justification of the results or a link to artifact}
      \\ \hline
      \small{...}
      & \small{...}
      & \small{...}
      \\ \hline
\end{tabular}
\\
\templatesection{Untested Requirements}
\begin{tabular}{|m{4cm}|m{12cm}|}
      \hline
      \rowcolor{Gray}
      \textbf{\small{Untested Requirement}}
      & \textbf{\small{Reason for not testing}} \\ \hline
      \rowcolor{white}
      \small{Requirement from requirements specification}
      & \small{Reason for not testing}
      \\ \hline
      \small{Requirement from requirements specification}
      & \small{Reason for not testing}
      \\ \hline
      \small{...}
      & \small{...}
      \\ \hline
\end{tabular}
 \\
\templatesection{Adversarial Testing}
\begin{tabular}{|m{4cm}|m{4cm}|m{8cm}|}
      \hline
      \rowcolor{Gray}
      \textbf{\small{Adversarial test}}
      & \textbf{\small{Results}}
      & \textbf{\small{Artifact}} \\ \hline
      \rowcolor{white}
      \small{Describe tests}
      & \small{Score or Results}
      & \small{Justification of the results or a link to artifact}
      \\ \hline
            \small{Describe tests}
      & \small{Score or Results}
      & \small{Justification of the results or a link to artifact}
      \\ \hline
            \small{...}
      & \small{...}
      & \small{...}
      \\ \hline
\end{tabular}
 \\
\templatesection{Other Testing }
\begin{tabular}{|m{4cm}|m{4cm}|m{8cm}|}
      \hline
      \rowcolor{Gray}
      \textbf{\small{Test}}
      & \textbf{\small{Results}}
      & \textbf{\small{Artifact}} \\ \hline
      \rowcolor{white}
      \small{Describe tests}
      & \small{Score or Results}
      & \small{Justification of the results or a link to artifact}
      \\ \hline
      \small{Describe tests}
      & \small{Score or Results}
      & \small{Justification of the results or a link to artifact}
      \\ \hline
      \small{...}
      & \small{...}
      & \small{...}
      \\ \hline
\end{tabular}
\vspace{4pt}
\\
\hline
\end{tabularx}
\end{table*}

\end{document}